\documentclass{article}
\usepackage{amsmath,graphicx,mlspconf,amssymb}
\usepackage{bm}
\usepackage{hyperref}
\hypersetup{
        colorlinks = true,
        allcolors = blue,
        breaklinks = true,
        linktocpage=true,
}
\usepackage{multirow}
\usepackage{booktabs}
\usepackage{balance}

\title{Closing the gap in multimodal medical representation alignment}
\name{Eleonora Grassucci, Giordano Cicchetti, Danilo Comminiello\thanks{This work was supported by the Italian Ministry of University and Research (MUR) within the PRIN 2022 Program for the project ``EXEGETE: Explainable Generative Deep Learning Methods for Medical Signal and Image Processing", under grant number 2022ENK9LS, CUP B53D23013030006. This work was also supported by the European Union under the National Plan for Complementary Investments to the Italian National Recovery and Resilience Plan (NRRP) of NextGenerationEU,  Project PNC 0000001 D3 4 Health, (Digital Driven Diagnostics, prognostics and therapeutics for sustainable Health care) - SPOKE 1 Clinical use cases and new models of care supported by AI/E-Health based solutions - CUP B53C22006120001, and under the NRRP of NextGenerationEU, partnership on “Future Artificial Intelligence Research” (PE00000013 – SPOKE 5 - CUP B53C22003980006 - FAIR: High Quality AI).}}
\address{Dept. of Information Engineering, Electronics, and Telecommunications\\Sapienza University of Rome, Italy}

\begin{document}

    

\maketitle

\begin{figure*}[ht]
    \centering
    \includegraphics[width=0.95\textwidth]{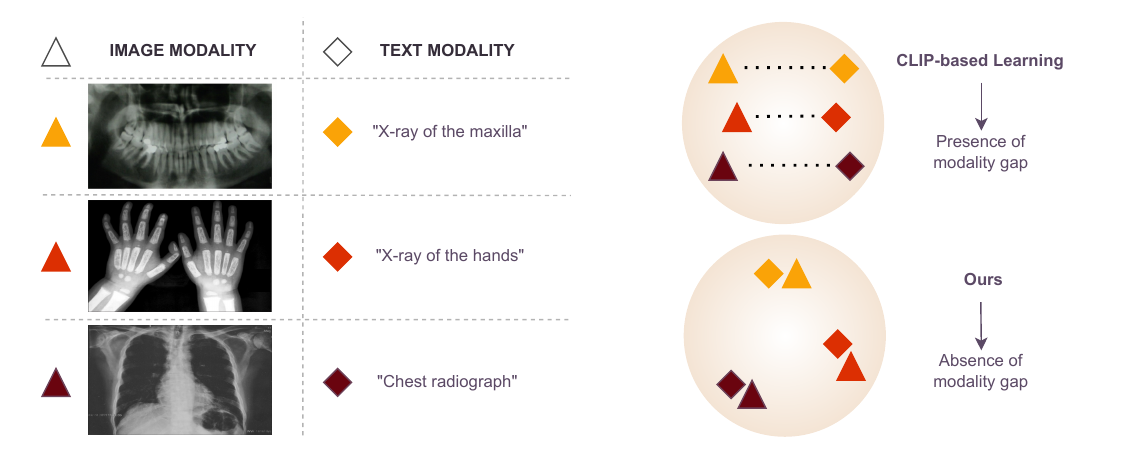}
    \caption{Illustration of the modality gap in the medical imaging domain. Triangles represent embeddings extracted by the image encoder, while diamonds represent embeddings from the text encoder. Colors indicate shared semantic meaning. In standard CLIP-based training, the resulting latent space shows a significant modality gap (i.e., embeddings from different modalities with the same meaning remain far apart). Our method introduces additional loss functions designed to reduce this gap and to align cross-modal embeddings closely based purely on the semantic meaning.}
    \label{fig:fig0}
\end{figure*}

\begin{abstract}

In multimodal learning, CLIP has emerged as the \textit{de facto} approach for mapping different modalities into a shared latent space by bringing semantically similar representations closer while pushing apart dissimilar ones. However, CLIP-based contrastive losses exhibit unintended behaviors that negatively impact true semantic alignment, leading to sparse and fragmented latent spaces. This phenomenon, known as the modality gap, has been partially mitigated for standard text and image pairs but remains unknown and unresolved in more complex multimodal settings, such as the medical domain.
In this work, we study this phenomenon in the latter case, revealing that the modality gap is present also in medical alignment, and we propose a modality-agnostic framework that closes this gap, ensuring that semantically related representations are more aligned, regardless of their source modality. Our method enhances alignment between radiology images and clinical text, improving cross-modal retrieval and image captioning.
\end{abstract}
\begin{keywords}
Multimodal Learning, Modality Gap, Medical Data Alignment
\end{keywords}

\begin{figure*}[t]
    \centering
    \includegraphics[width=0.8\textwidth]{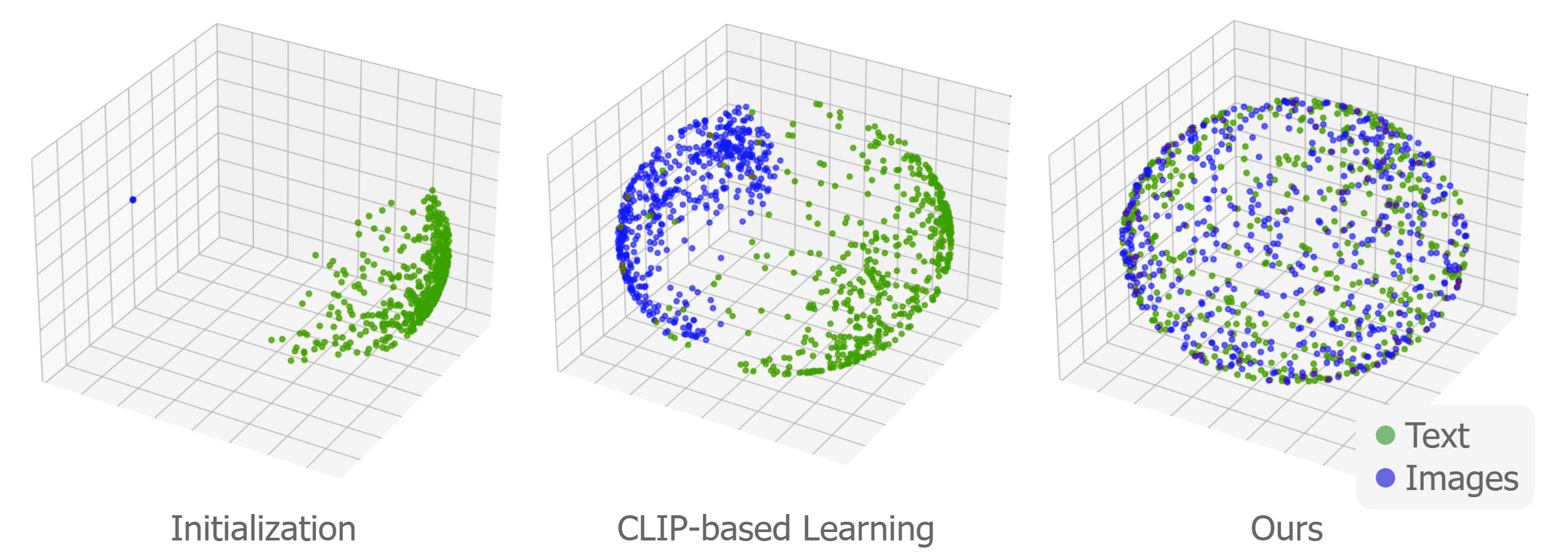}
    \caption{The modality gap originates at initialization, with the two modalities clearly clustered. Such a gap still persists even after training in the conventional CLIP-based learning setting, such as MedCLIP \cite{Wang2022MedCLIPCL}. On the contrary, the proposed method closes the gap, leveraging the whole space to cluster embeddings according to the semantics rather than to the modality type.}
    \label{fig:fig1}
\end{figure*}

\section{Introduction}
\label{sec:intro}

Multimodal representation learning has emerged as a fundamental paradigm in machine learning, enabling models to process and integrate data from multiple sources such as text, images, and audio. The core assumption of multimodal learning is that representations with similar semantics, regardless of their source modality, should be mapped close to one another within a shared latent space. However, despite the success of contrastive learning approaches like CLIP \cite{Radford2021LearningTV}, a persistent issue prevents this ideal alignment: the modality gap \cite{liang2022mind}. When multimodal data are projected into a shared latent space, samples from the same modality tend to initially cluster together, forming distinct modality-specific groupings. Unfortunately, these clusters persist even after training, resulting in a sparse and fragmented latent space \cite{Eslami2024MitigateTG, Fahim2024ItsNA}. Here, certain modalities are densely packed into small regions, while others are more widely distributed, as shown in Fig.~\ref{fig:fig1}. After convergence in conventional CLIP-based training, embeddings may appear aligned yet remain distant in the latent space, as matching pairs tend to cluster according to their modality rather than forming semantically coherent groups. This occurs because representations from different modalities tend to form separate clusters, disrupting semantic coherence and significantly impairing downstream performance. 


Concurrently, multimodal learning has seen growing adoption in the medical domain, where integrating multiple sources of information, such as radiology images and clinical text,
has the potential to improve diagnostic accuracy and clinical decision-making \cite{Wang2022MedCLIPCL, zhang2023biomedclip, chaves2024clinically, Kumar2024ImprovingMM}.
However, despite the advancements in multimodal learning, to the best of our knowledge, no prior studies have investigated the impact of the modality gap on medical data
alignment. The modality gap may introduce severe limitations when learning representations from heterogeneous and content-rich medical data.
Indeed, if data is not properly aligned, tasks such as cross-modal retrieval, or captioning may suffer from reduced accuracy and reliability
\cite{yaras2024explainingmitigatingmodalitygap}. Interestingly, in this paper, we find that the modality gap exists in medical data too, and more seriously, that true pairs are poorly aligned.
On average, with the conventional CLIP loss function, true pairs have indeed a cosine similarity of 0.20, corresponding to an angle of 80 degrees.
In practice, true pairs are almost orthogonal in the multimodal latent space. Therefore, it is crucial to better align true pairs, close the gap, and build a more aligned embedding space to represent multimodal medical data. Indirectly, a model that inconsistently aligns different modalities or fails to provide coherent predictions across imaging and textual data may crucially undermine the confidence of clinicians in AI-assisted diagnostic tools.

Following our exploration of the impact of the modality gap on multimodal medical data, we propose a novel approach that closes the modality gap across the modalities, enabling the creation of a truly unified and structured multimodal latent space. In the experimental evaluation, we show that the proposed method better aligns matching pairs in the latent space and is effectively capable of closing the modality gap, enhancing the performance in downstream tasks.

The rest of the paper is organized as follows. In Section \ref{sec:work} we explore the related work, in Section \ref{sec:method} we propose the method to close the gap, while Section \ref{sec:exp} shows the experimental evidences and Section \ref{sec:con} draws the conclusions.


\section{Related Work}
\label{sec:work}

\textbf{Multimodal Learning.} Starting from the cosine similarity-based CLIP losses \cite{Radford2021LearningTV} several multimodal models have been developed for
two modalities like CLAP \cite{CLAP2022} or CLIP4Clip \cite{Luo2021CLIP4ClipAE}. Lately, the same loss has been extended to multiple modalities in ImageBind
\cite{Girdhar2023ImageBindOE}, LanguageBind \cite{Zhu2023LanguageBindEV}, or VAST \cite{Chen2023VASTAV}. More recently, novel approaches have been proposed
for multimodal learning to avoid the cosine similarity loss and rethinking multimodal alignment, namely GRAM \cite{cicchetti2024gram} that relies on the volume computation and Symile \cite{saporta2024contrasting}, based on total correlation. In the medical domain, the widespread method to align pairs of images and text is MedCLIP \cite{Wang2022MedCLIPCL}, which still relies on the conventional CLIP-based loss function.

\textbf{Modality Gap.} The modality gap has been observed for the first time by \cite{liang2022mind}, and then studied mainly for the CLIP model
\cite{wu2023understanding, Fahim2024ItsNA,shi2023towards} or for generic image and text pairs \cite{yaras2024explainingmitigatingmodalitygap, iclr2024two, wu2023understanding}.
These works provide theoretical justification for the gap and propose to mitigate the gap by fixing the temperature
\cite{yaras2024explainingmitigatingmodalitygap}, by applying post-hoc translations in the latent space \cite{liang2022mind, iclr2024two},
or by sharing the transformer encoder and the projection layer in the vision and language encoders \cite{Eslami2024MitigateTG}.
Overall, these methods studied the modality gap in the case of generic image-text pairs, without advancing clues on the case of specific kinds of data with information imbalance, such as medical data.

\section{Proposed Method}
\label{sec:method}

In this Section, we present the limitations of current multimodal models and the proposed novel solution. In particular, we define two novel loss functions that can close the modality gap and better align matching pairs in the latent space.

\textbf{Notation.} Given a set of $M$ modalities, $M_i$ is the $i$-th modality (i.e., text), $m_i$ is a sample from the modality $M_i$ (i.e., \textit{``A radiology of the left hand"}), while $\mathbf{m}_i$ is the learned latent representation of the sample $m_i$.

\subsection{What is the Modality Gap?}
Suppose to have a set of $n$ samples from $M$ modalities, with $\{ (m_{i, 1}, m_{i, 2}, \dots, m_{i, M}\}_{i=1}^n$ being paired samples from the $M$ modalities. That is, $m_{i, 1}$ is the radiology of a hand, $m_{i, 2}$ and the caption \textit{``A radiology of a hand"}. The scope of multimodal learning is to train modality-specific encoders $e_M : m_M \rightarrow \mathbf{m}_M, \mathbf{m}_M \in \mathrm{R}^d $ that map the data into a shared multimodal latent $d$-dimensional space where representations with similar semantics cluster close to each other even though they come from different modalities. 
To this end, a huge number of contrastive loss functions have been designed, among which the most common is the one introduced in CLIP \cite{Radford2021LearningTV} for text and image modalities, recall also from MedCLIP \cite{zhang2023biomedclip} for medical text-image alignment.
The conventional contrastive loss function is expressed in terms of cross-entropy objective:
\begin{align}
        \mathcal{L}_{M_1 \rightarrow M_2}&=-\frac{1}{B}\sum_{i=1}^{B}\log\frac{\exp(\textbf{m}_{1,i}^\top \textbf{m}_{2,i}/\tau)}{\sum_{j=1}^{B}\exp(\textbf{m}_{1,i}^\top \textbf{m}_{2,j}/\tau)}\\
    \mathcal{L}_{M_1 \rightarrow M_2} &= \frac{1}{B} \sum_{i=0}^B H(\mathbb{1}_i,p_i),
\label{eq:entropy}
\end{align}
in which $\tau$ is the temperature parameter, $H$ is the cross-entropy function in charge of aligning the one-hot distribution $\mathbb{1}_i$ to the probability density function $p_{i}$, whose elements correspond to:
\begin{equation}
    p_{i,j} = \frac{\exp(\textbf{m}_{1,i}^\top \textbf{m}_{2,i}/\tau)}{\sum_{j=1}^{B}\exp(\textbf{m}_{1,i}^\top \textbf{m}_{2,j}/\tau)}.
\end{equation}
To obtain the final CLIP loss usually works average the two directions of the objectives, $\mathcal{L}_{M_1 \rightarrow M_2}$ and $\mathcal{L}_{M_2 \rightarrow M_1}$, to account for the non-symmetry of the cross-entropy.

The gap between modalities emerges during the initialization phase, where encoders initialized with random weights project data into distinct, narrow cones within the shared latent space \cite{liang2022mind}, as illustrated in Fig. \ref{fig:fig1}. This initial gap persists throughout training. Consequently, even though the final learned representations achieve some semantic alignment, positive pairs remain disconnected and significantly separated, as depicted in the CLIP-based learning plot in Fig. \ref{fig:fig1}. Prior studies \cite{shi2023towards, cicchetti2024gram} have shown that the conventional CLIP loss function often gets trapped in local minima, producing partially matched yet distant positive pairs.
Previous research also reveals that the CLIP loss comprises two components with distinct objectives \cite{Eslami2024MitigateTG}: one encourages alignment of positive pairs, while the other promotes separation of non-matching pairs. In practice, these opposing forces often balance each other, preventing the modality gap from closing. Instead, training leads to modality representations aligning in “semantic stripes,” as illustrated on the left-hand side of Fig.~\ref{fig:fig1}. Despite this imperfect alignment, characterized by positive pairs having cosine similarities significantly lower than the ideal value of 1.0, these semantic stripes still enable effective retrieval performance because positive pairs retain higher similarity relative to non-matching pairs.
Therefore, the representations of true matching pairs remain relatively distant in the latent space, failing to occupy the same region. This separation limits both the expressiveness and the semantic alignment of the multimodal latent space.





\subsection{Closing the Modality Gap}
Our objective is to close the modality gap while maintaining consistent alignment across the distribution of positive pairs. To this end, we introduce two novel loss functions. The first, called the Align True Pairs loss ($\mathcal{L}_{\text{ATP}}$), enforces alignment specifically between true positive pairs. We consider a scenario involving $M$ modalities, where one modality, denoted as $\mathbf{a}$, serves as the anchor, i.e., the reference modality to which all other modalities are aligned \cite{Girdhar2023ImageBindOE}. The loss is defined as:
\begin{equation}
    \mathcal{L}_{\text{ATP}} =  \frac{1}{M-1} \sum_{i=0, \mathbf{m}_i \neq \mathbf{a}}^{M} \left( \frac{1}{B} \sum_{j=0}^B \left( || \mathbf{m}_{ij} - \mathbf{a}_j ||^2_2 \right) \right), 
\end{equation}
where $m_i$ is the $i$-th modality and $B$ the batch size.
The second one, the Centroid Uniformity loss $\mathcal{L}_{\text{CU}}$, ensures uniformity among the modalities in the latent space by:
\begin{equation}
    \mathcal{L}_{\text{CU}} = \log \left( \frac{1}{B} \sum_{i=0}^{B} \sum_{j=0, j\neq i}^{B} \exp \left( -2 || \mathbf{c}_i - \mathbf{c}_j ||^2_2 \right) \right), 
\label{eq:unif}
\end{equation}
in which $\mathbf{c}_k$, with $k = {i,j}$, are the centroids defined as:
\begin{equation}
    \mathbf{c}_k = \frac{1}{M} \sum_{k=0}^{M} \mathbf{m}_k,
\end{equation}
and $\mathbf{c}_k$ is the centroid of the $k$-th element of the batch built by averaging all the modality embeddings.
The effect of the two losses is complementary. On one side, the $\mathcal{L}_{\text{ATP}}$ enhances closeness between positive pairs, effectively enhancing the mean cosine similarity between them. 
However, relying solely on this loss can lead to an undesirable outcome: the collapse of the latent space into small regions, causing representations of semantically unrelated data to overlap. To address this, the contribution of the $\mathcal{L}_{\text{CU}}$ loss is essential, as it promotes sparsity in the latent space by enforcing uniform distribution of the centroids while preserving the established alignment. Adjusting the positions of centroids naturally shifts the associated modality representations, thereby maintaining alignment and utilizing the full extent of the latent space. In the absence of centroids, enforcing uniformity separately for each modality would risk disrupting the alignment between semantically related pairs. Furthermore, the radial basis function (RBF) kernel used in Equation~\eqref{eq:unif} aligns well with the concept of a uniform distribution over the unit hypersphere, where multimodal embeddings typically reside \cite{wang2020understanding}, thus encouraging coverage across the entire surface. Importantly, since the method operates at the loss level, it remains modality-agnostic and applies to any type of modality.
%
The final proposed loss function, whose aim is to jointly align the true pairs and close the modality gap, is a sum of the two terms:
\begin{equation}
    \mathcal{L}_{\text{gap}} = \mathcal{L}_{\text{ATP}} + \mathcal{L}_{\text{CU}}.
\end{equation}
Such loss should then be combined with the contrastive loss to obtain:
\begin{equation}
    \mathcal{L}_{\text{CL}_{\text{gap}}} = \mathcal{L}_{\text{gap}} + \frac{1}{2} \left( \mathcal{L}_{M_1 \rightarrow M_2} + \mathcal{L}_{M_2 \rightarrow M_1} \right).
\label{eq:loss_gap}
\end{equation}

\begin{table*}
    \centering
    \caption{Latent space alignment (Cos True Pairs and Gap) and retrieval results (Recall at 1,5,10) on the ROCO dataset.}
    \begin{tabular}{l|cc|ccc}
        \toprule
        Method & Cos True Pairs $\uparrow$ & Gap $\downarrow$ & R@1 & R@5 & R@10  \\ \midrule
        CLIP (LT) \cite{Radford2021LearningTV} & 0.20 & 0.40 & \textbf{39.5} & 67.4 & 74.4  \\
        CLIP (FT) \cite{yaras2024explainingmitigatingmodalitygap} & 0.39 & 0.14 & 38.3 & 65.8 & 75.8  \\
        Ours & \textbf{0.54} & \textbf{0.12} & 38.9 & \textbf{68.8} & \textbf{81.8} \\
        \bottomrule
    \end{tabular}
    \label{tab:Med_retrieval}
\end{table*}

\begin{table*}
    \centering
    \caption{Captioning results on the ROCO dataset.}
    \begin{tabular}{l|cccccc}
        \toprule
        Method & BLEU@1 $\uparrow$ & BLEU@2 $\uparrow$ & BLEU@3 $\uparrow$ & BLEU@4  $\uparrow$ & ROUGE L $\uparrow$ & CIDER $\uparrow$\\ \midrule
        CLIP (LT) \cite{Radford2021LearningTV} & 16.51 & 9.82 & 5.91 & 3.56 & 19.61 & 25.24 \\
        CLIP (FT) \cite{yaras2024explainingmitigatingmodalitygap} & 16.71 & \textbf{10.07}  & 6.05 & 3.59 & 19.82 &\textbf{26.05}\\
        Ours & \textbf{16.96}  & \textbf{10.07} & \textbf{6.09} & \textbf{3.64} & \textbf{19.90}& 25.22 \\
        \bottomrule
    \end{tabular}
    \label{tab:Med_Captioning}
\end{table*}

\subsection{Measuring the Latent Space Alignment}
\label{sec:metrics}

Following \cite{liang2022mind}, to measure such a gap between two generic modalities $M_i$ and $M_j$, we measure the effective Euclidean distance between the centroids of each modality:
\begin{equation}
    \text{Gap}_{M_i, M_j} = \| \textbf{c}_{m_i} - \textbf{c}_{m_j} \|,
\end{equation}
where $\textbf{c}_{m_i} = \sum_{\textbf{m} \in M_i}\textbf{m}$. Even though the gap is zero, this does not imply that the embeddings are effectively aligned in the latent space. Therefore, we further evaluate the mean cosine similarity true pairs metric, defined as: 
\begin{equation}
    \text{Cos True Pairs}_{M_i, M_j} = \frac{1}{N} \sum_{k=0}^N  \left( \textbf{m}_{i,k} \cdot \textbf{m}_{j,k}^\top \right).  
\end{equation}
This metric quantifies the proximity of normalized matching pairs within the latent space. A value approaching 1.0 indicates a smaller angle between the vectors, signifying that the matching pairs are closely positioned. Additionally, we introduce the mean angular value (AV) metric, which is computed within each modality. Its formulation is given by:

\begin{equation}
    \text{AV}_M = \frac{1}{N^2-N}\sum_{i=0}^N\sum_{j=0, j\neq i}^N \left( \textbf{m}_i \cdot \textbf{m}_j^T \right).
\end{equation}

This metric captures the average cosine similarity within a single modality, providing insight into how the embeddings are distributed across the hypersphere. A value close to 1.0 suggests that the embeddings are tightly clustered, whereas a value near 0 indicates a wide dispersion, with cosine similarities spanning from -1 to 1, signifying effective sparsification of the latent space.

\section{Experimental Results}
\label{sec:exp}

\textbf{Settings.} To perform the study on how the modality gap impacts multimodal medical data, we involve the Radiology Object in Context (ROCO) dataset \cite{ROCOdataset}, containing two modalities. The dataset comprises images from publications available on the PubMed Central Open Access FTP mirror, which were automatically detected as non-compound and either radiology or non-radiology. In this scenario we select only the radiology set comprising of 65420 images for training and 8176 for testing. Each image (no specific body region is selected) is associated with a caption. Captions are very heterogeneous, and comprise examples like \textit{"Showing the subtrochanteric fracture in the porotic bone."} up to like \textit{"A 3-year-old child with visual difficulties. Axial FLAIR image shows a supra-sellar lesion extending to the temporal lobes along the optic tracts (arrows) with moderate mass effect, compatible with optic glioma. FLAIR hyperintensity is also noted in the left mesencephalon from additional tumoral involvement."}.
To process such data and obtain embeddings we select larger models with respect to the ones in the previous section. As image encoder, we involve EVAClip-ViT-G ($\sim$ 1B parameters), which shows improved performance in zero-shot multimodal scenarios \cite{Sun2023EVACLIPIT}. Images are randomly cropped to the fixed dimension 224x224. Overall, the vision encoder could be seen as a mathematical nonlinear function that maps images from their pixel space $\mathcal{R}^{hxw}$ to a latent vectorial representation $\mathcal{R}^d$, i.e., $e_I: \mathcal{R}^{hxw} \rightarrow \mathcal{R}^d$.

\noindent As text encoder, we involve BERT-B. The text encoder maps the text in input, modelled as a sequence of tokens, into a vectorial representation $\mathcal{R}^d$, i.e., $e_T: \mathcal{R}^{t,t_{\text{dim}}} \rightarrow  \mathcal{R}^d$, where $t$ is the number of tokens for input text and $t_{\text{dim}}$ is the dimension in which the text tokenizer represents the tokens. In our case, $t=70$ and $t_{\text{dim}} = 1$. Padding to 0 is used if needed.

These two encoder models are already proven to be effective in processing multimodal data \cite{Chen2023VASTAV}. Both encoders are transformer-based encoders. We select the classification token, CLS, as output latent embedding. The dimensionality $d$ of the shared latent space is set to 512.
 
We conduct retrieval experiments using the conventional CLIP loss function with the learnable temperature parameter \cite{Radford2021LearningTV}, as the well-known MedCLIP. Then, as suggested by  \cite{yaras2024explainingmitigatingmodalitygap}, we fixed the temperature to $0.07$ allowing for a partial gap reduction. The last experiments is done using our proposed $\mathcal{L}_{\text{CL}_{\text{gap}}}$.
In all the experiments we train the aforementioned framework for 100 epochs using AdamW as optimizer with a fixed learning rate of $1e-4$.

We perform the common image-text retrieval task. Moreover, to further investigate the effectiveness of the proposed loss function in downstream tasks we perform the image captioning task. Following \cite{yan2022videococa}, we involve a decoder serving as the language generator model and we add a specific loss term that, along with our proposed loss functions, trains the text encoder-decoder structure for this specific task. Intuitively, the more aligned the latent space, the better the model will generate the captions from the latent representations. The captioning loss function is:
\begin{equation}
\label{eq:ldam}
    \mathcal{L}_{\text{cap}}= - \sum_{t=1}^T \log P_\theta(\mathbf{y}_t|\mathbf{y}_{<t}, \mathbf{x}),
\end{equation}
where $\mathbf{y}$ is the exact tokenized texts the model aims to learn by maximizing the conditional likelihood under the forward autoregressive factorization. $P_\theta(\mathbf{y}_t|\mathbf{y}_{<t}, \mathbf{x})$ denotes the probability assigned to the token $\mathbf{y}_t$ given as input the past history $\mathbf{y}_{<t}$ and the input features $\mathbf{x}$.

\textbf{Metrics.} Along with standard metrics to measure the semantic alignment of the latent space (i.e., Cos True Pairs, Gap, and Angular Value) introduced in Sec.~\ref{sec:metrics}, we evaluate the performance in downstream tasks. For the retrieval task, we involve the Recall @1,5,10. For the image captioning task, we consider standard metrics such as BLEU@1, BLEU@2, BLEU@3, BLEU@4, ROUGE-L, and CIDEr.

\textbf{Results.} Tab.~\ref{tab:Med_retrieval} and Tab.~\ref{tab:Med_Captioning} show the results on the correlation between gap reduction and performance gain. According to the scores in Tab.~\ref{tab:Med_retrieval}, our introduced loss function can reduce the gap between image and text modalities down to 0.12 while ensuring the closeness of true pairs in the latent space up to 0.54. This result is also evident in Fig. \ref{fig:fig1}, in which our method better spreads over the whole latent space by closing the modality gap. Interestingly, building such a more aligned space crucially improves the retrieval performance, especially in R@10. The latter result is important, as R@10 measures if the correct result is within the first ten samples in descending cosine similarity order. Our methods clearly improve this metric up to 7.4 points. This means that the latent space is overall better aligned and that is more likely that the model places the correct data close to the query.
Captioning results are instead shown in Tab.~\ref{tab:Med_Captioning}, and they prove our intuition on the importance of the aligned latent space. According to all the metrics, shaping a better-aligned multimodal latent space considerably enhances the decoder performance in generating the image caption with respect to conventional methods.

\section{Conclusion}
\label{sec:con}

In this paper, we addressed the critical issue of the modality gap within multimodal learning frameworks, particularly in the context of medical data alignment. We demonstrated that conventional contrastive approaches, such as CLIP-based models, result in suboptimal semantic alignment between radiology images and clinical text, negatively affecting downstream tasks like cross-modal retrieval and image captioning. To overcome this limitation, we introduced a novel approach featuring two complementary losses, the Align True Pairs loss and the Centroid Uniformity loss. Our proposed method effectively closes the modality gap by significantly improving the proximity of semantically similar pairs within the latent space, thereby enhancing semantic coherence.

Experimental validation on the ROCO dataset confirmed that our framework substantially reduces the modality gap and achieves superior performance in downstream tasks. Specifically, we observed notable improvements in retrieval accuracy, especially in Recall@10, highlighting the improved alignment of matching pairs. Additionally, our approach enhanced the quality of generated captions, demonstrating broader applicability and efficacy in multimodal medical scenarios. Future research directions include extending our methodology to additional modalities and investigating its impact on real-world clinical applications.

\ninept
\bibliographystyle{IEEEbib}
\bibliography{biblio}
\balance

\end{document}